\documentclass[conference]{IEEEtran}
\IEEEoverridecommandlockouts
\usepackage{cite}
\usepackage{amsmath,amssymb,amsfonts}
\usepackage{algorithmic}
\usepackage{graphicx}
\usepackage{textcomp}
\usepackage{microtype}
\usepackage{multirow,array,booktabs}
\usepackage{placeins}
\usepackage{float}
\usepackage[math]{cellspace}
\usepackage{caption}
\usepackage{subcaption}
\usepackage{booktabs} % for professional tables
\usepackage[table]{xcolor}
\newcommand\y{\cellcolor{blue!10}}
\newcommand\yy{\cellcolor{green!10}}
\newcommand\yyy{\cellcolor{red!10}}
\def\BibTeX{{\rm B\kern-.05em{\sc i\kern-.025em b}\kern-.08em
    T\kern-.1667em\lower.7ex\hbox{E}\kern-.125emX}}

\makeatletter
\newcommand{\linebreakand}{%
  \end{@IEEEauthorhalign}
  \hfill\mbox{}\par
  \mbox{}\hfill\begin{@IEEEauthorhalign}
}
\makeatother
    
\begin{document}

\title{Anomaly Attribution of Multivariate Time Series using Counterfactual Reasoning}

% \author{\IEEEauthorblockN{1\textsuperscript{st} Violeta Teodora }
% \IEEEauthorblockA{\textit{Anonymous Department} \\
% \textit{Anonymous Institution}\\
% City, Country \\
% anonymous.author@anonymous.an}}
\author{\IEEEauthorblockN{1\textsuperscript{st} Violeta Teodora Trifunov}
\IEEEauthorblockA{\textit{Computer Vision Group} \\
\textit{Friedrich Schiller University Jena}\\
Jena, Germany \\
violetateodora.trifunov@uni-jena.de}
\and
\IEEEauthorblockN{2\textsuperscript{nd} Maha Shadaydeh}
\IEEEauthorblockA{\textit{Computer Vision Group} \\
\textit{Friedrich Schiller University Jena}\\
Jena, Germany \\
maha.shadaydeh@uni-jena.de}
\and
\IEEEauthorblockN{3\textsuperscript{rd} Bj\"orn Barz}
\IEEEauthorblockA{\textit{Computer Vision Group} \\
\textit{Friedrich Schiller University Jena}\\
Jena, Germany \\
bjoern.barz@uni-jena.de}
\linebreakand
\IEEEauthorblockN{4\textsuperscript{th} Joachim Denzler}
\IEEEauthorblockA{\textit{Computer Vision Group} \\
\textit{Friedrich Schiller University Jena}\\
\textit{Michael Stifel Center Jena for Data-Driven and Simulation Science}\\
\textit{German Aerospace Center (DLR), Institute for Data Science}\\
Jena, Germany}
}

\maketitle

\begin{abstract}
There are numerous methods for detecting anomalies in time series, but that is only the first step to understanding them. We strive to exceed this by explaining those anomalies. Thus we develop a novel attribution scheme for multivariate time series relying on counterfactual reasoning. We aim to answer the counterfactual question of would the anomalous event have occurred if the subset of the involved variables had been more similarly distributed to the data outside of the anomalous interval. Specifically, we detect anomalous intervals using the Maximally Divergent Interval (MDI) algorithm, replace a subset of variables with their in-distribution values within the detected interval and observe if the interval has become less anomalous, by re-scoring it with MDI. We evaluate our method on multivariate temporal and spatio-temporal data and confirm the accuracy of our anomaly attribution of multiple well-understood extreme climate events such as heatwaves and hurricanes.
\end{abstract}

\begin{IEEEkeywords}
anomaly attribution, multivariate time series, counterfactual reasoning
\end{IEEEkeywords}

\section{Introduction}\label{intro}
\begin{figure}
    \centering
    \includegraphics[width=1\columnwidth]{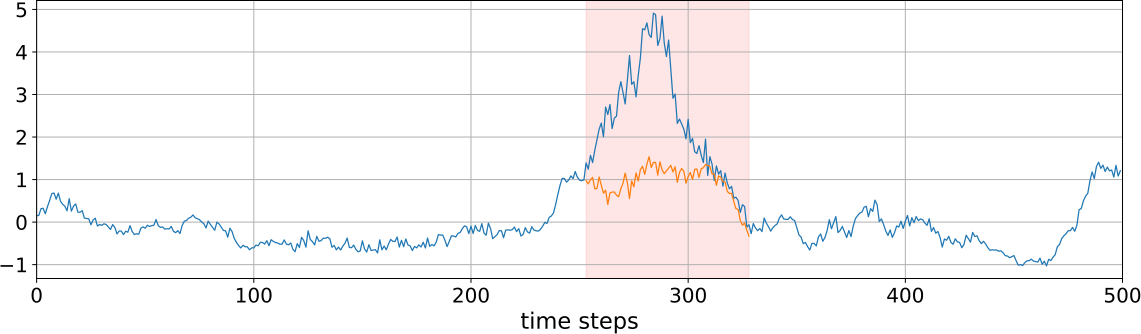}
    \caption{Counterfactual replacement of the wave height (Hs) variable (orange) within the anomalous interval highlighted in red and the data before the replacement (blue).}
    \label{fig:time-dependent-replacement}
\end{figure}

Finding causes of extreme weather events, power outages and abnormal fluctuations in financial data can be of crucial importance for their understanding and taking precautionary measures or even preventing them from occurring again in the future. We propose a novel anomaly attribution scheme to analyze anomalous intervals of multivariate temporal and spatio-temporal data and attribute those anomalies to a set of involved variables. To achieve this, we engage in answering the counterfactual question: would the anomalous event have occurred if one or more of the involved variables had been more similarly distributed to the data outside of the anomalous interval. For example, would there have been a heatwave in Europe, had the air temperature been lower in the summer that year or more similar to the last year's summer's average. However, since this question cannot be answered directly, we do the next best thing and replace the subset of variables in the detected anomalous interval by a random sample from their nominal distribution, as illustrated in Fig. \ref{fig:time-dependent-replacement} for a single variable. Our novel replacement algorithm preserves both the inter-variable and the inter-temporal correlations of the data.

We then examine whether the same interval became less anomalous after replacing a specific subset of variables by applying the Maximally Divergent Interval (MDI) method \cite{MDI} for anomaly detection, repeatedly on each of the subsets. By determining which variables yield the lowest anomaly score after the replacement, we can conclude that the subset of variables in question was the reason why the anomaly had occurred.

Our attribution method can be applied to any multivariate time series data regardless of potential outliers and missing values. This also makes it appropriate for the task of data repairing. Furthermore, it is one of the few attribution schemes for multivariate time series, and offers interpretation of the MDI's anomaly detections.

The remainder of the paper is organized as follows. We discuss other time series anomaly attribution approaches in Section \ref{related_work}. In Section \ref{background}, we introduce the basic concepts of causality and counterfactual reasoning and briefly review the MDI algorithm for anomaly detection. Our novel anomaly attribution approach for multivariate time series is introduced in Section \ref{method}. We evaluate our method by attributing anomalies on multiple datasets of ecological time series and the spatio-temporal climate data where the causes of the anomalies are well-understood. The results are presented in Section \ref{experiments} and Section \ref{conclusion} concludes this paper.

\section{Related Work}\label{related_work}

With the increase of the available data, creating efficient methods which can analyze it, detecting anomalies, and explaining what caused the anomalies as reliably as possible becomes crucial. There are numerous anomaly detection approaches for time series \cite{anomaly-detection-review}, but not many are concerned with attributing those anomalies. One of them, however, is the work by Siddiqui et al. \cite{TSInsight}, where the authors introduce a time series anomaly attribution method called TSInsight, which uses an auto-encoder with a sparsity-inducing norm on its output to the classifier. It learns to keep the features that are important for prediction by the classifier and discards the unimportant ones. TSInsight can create both instance-based and model-based explanations. This feature attribution method, unlike our approach, does not use counterfactual variables to attribute anomalies. Moreover, it does not attribute the anomalies to variables in the input data directly but to features derived from them. However, similar to our idea, Id\'e et al. \cite{LC} propose to explain the anomalous prediction produced by a black-box regression model by inferring the responsibility score of each of the input variables. They introduce the Likelihood Compensation (LC) method, based on the likelihood principle, i.e., on the proposition that, given a statistical model, all the evidence in a sample that is pertinent to the model parameters is enclosed in the likelihood function. This method computes a correction to each input variable. In contrast to our approach, however, the LC method does not consider replacing certain intervals of a subset of variables and is only applied to anomalous forecasts. Using MDI, we can detect any anomalous interval in the entire time series and correct and attribute it using our method, regardless of any outliers or missing values. Next, Zhang et al. \cite{ts-repair} also engage in iterative repairing of anomalous time series data for improving pattern mining or classification tasks. Unlike our method, it does not aim to provide answers to what caused the anomalies themselves and is more computationally intensive. In the work of Shadaydeh et al. \cite{Maha-GCPR}, anomalies are attributed based on the changes in the spectral cause-effect relationships between each pair of the involved variables. Unlike our method, it does not identify the subset of variables that contribute most to the anomalous event. Furthermore, in the method for attribution of multivariate extreme events proposed by Guanche et al. \cite{Maha-CI}, the authors aim to answer the question of how much does each variable contribute to the Mahalanobis distance metric. However, this method is applied on each time step independently rather than interval-wise as is the case with our method.

\section{Methodological Background}\label{background}

We will now briefly discuss the fundamental concepts of causality and counterfactual inference, followed by a more in-depth explanation of the method we use for detecting the anomalous intervals of multivariate time series to which our novel attribution scheme is tailored.

\subsection{Causal and Counterfactual Reasoning}

To better understand the world, humans seek to uncover the causes of certain events or phenomena by observing them over time. However, for different reasons, the answer to the question of why something happened is sometimes difficult to obtain. One way to empirically determine a cause of a particular event is to alter one of its potential causes at a time and observe if the alteration changed the event in question. This alteration is formally known as \textit{intervention}. The rules governing the interventions are defined by the \textit{do}-calculus \cite{Pearl}. However, when the outcome of an event of interest cannot be reproduced due to the intervention being impractical or unethical, one needs to turn to the \textit{counterfactual reasoning}. Namely, one can perform a thought experiment, asking oneself how the world would have changed had certain different actions been taken. In our case, it is impractical to intervene on climate variables and observe what kind of intervention would make a specific event less anomalous. Still, we can answer the question of how the anomalous event would have changed had a specific subset of the involved variables been distributed like the rest of the data outside of the anomaly. When this type of intervention results in a less anomalous event, we attribute it to the subset of the variables on which we intervened.

\subsection{MDI Algorithm for Anomaly Detection}

The MDI \cite{MDI} is an unsupervised method for detecting anomalies in multivariate temporal and spatio-temporal data. Following the notation by Barz et al. \cite{MDI}, we let $X := \{x_{t}\}^{n}_{t=1}$ be a time series for $x_{t} = [ x^{1}_{t}, \dots, x^{d}_{t} ]^{\top} \in \mathbb{R}^{d}$, and $n, d \in \mathbb{N}$. By $x^{j}:=\{x^{j}_{t}\}_{t=1}^{n}$, for $j \in \mathcal{J} := \{1, \dots, d\}$, we denote each of the $d$ variables constituting $\{x_{t}\}^{n}_{t=1}$. The set of all subintervals of the set $\{1, \dots, n\}$ is denoted by $\mathcal{I}$. To find the most anomalous interval $I = \{t \in \mathbb{N} \ | \ a \leq t < b\} = [a, b) \cap \mathbb{N} \in \mathcal{I}$, for $a, b \in \{1, \dots, n\}$, of the time series $\{x_{t}\}^{n}_{t=1}$, the MDI algorithm looks for an interval whose data distribution $p_{I}$ is the most different from the distribution of the data in the rest of the time series with indices in $\Omega = \{1,\dots, n\} \setminus I$, denoted by $p_{\Omega}$. This difference of the distributions $p_{I}$ and $p_{\Omega}$ is quantified using the Kullback-Leibler (KL) divergence:
\begin{equation}\label{eq:KL-div}
    \mathcal{D}(p_{I}, p_{\Omega}) = \int p_{I}(x_{t}) \cdot \log \left( \frac{p_{I}(x_{t})}{p_{\Omega}(x_{t})}\right) dx_{t}.
\end{equation}
\begin{figure*}[h!tb]
    \fcolorbox[gray]{0.75}{0.99}{%
    \begin{minipage}[t]{.99\linewidth}
    \centering
    \resizebox{.95\linewidth}{!}{%
    \begin{minipage}[t]{\linewidth}
    \begin{align*}\label{eq:cov-matrix}
        \begin{split}
            \textrm{cov}(X') &= \mathbb{E}_{t}(x'_{t} \cdot {x'_{t}}^{\top}) = \mathbb{E}_{t}\left(
            \begin{bmatrix}
                x_{t}\\
                x_{t-1}\\
                x_{t-2}
            \end{bmatrix} 
            \cdot
            \begin{bmatrix}
                x_{t}^{\top} & x_{t-1}^{\top} & x_{t-2}^{\top}
            \end{bmatrix} \right)
            =\mathbb{E}_{t}\left(
            \begin{bmatrix}
                x^{1}_{t}\\
                x^{2}_{t}\\
                x^{1}_{t-1}\\
                x^{2}_{t-1}\\
                x^{1}_{t-2}\\
                x^{2}_{t-2}\\
            \end{bmatrix}
            \cdot
            \begin{bmatrix}
                {x^{1}_{t}} & {x^{2}_{t}} & {x^{1}_{t-1}} & {x^{2}_{t-1}} & {x^{1}_{t-2}} & {x^{2}_{t-2}}
            \end{bmatrix} \right)\\
            &=\mathbb{E}_{t}\left(
            % \begin{bmatrix}
            \renewcommand{\arraystretch}{1.5}
            \left[
            \begin{array}{c c|cc|cc}
                 \y x^{1}_{t} {x^{1}_{t}} & \y x^{1}_{t} {x^{2}_{t}} & \yy x^{1}_{t} {x^{1}_{t-1}} & \yy x^{1}_{t} {x^{2}_{t-1}} & \yyy x^{1}_{t} {x^{1}_{t-2}} & \yyy x^{1}_{t} {x^{2}_{t-2}}\\
                 \y x^{2}_{t} {x^{1}_{t}} & \y x^{2}_{t} {x^{2}_{t}} & \yy x^{2}_{t} {x^{1}_{t-1}} & \yy x^{2}_{t} {x^{2}_{t-1}} & \yyy x^{2}_{t} {x^{1}_{t-2}} & \yyy x^{2}_{t} {x^{2}_{t-2}}\\ \hline
                 \yy x^{1}_{t-1} {x^{1}_{t}} & \yy x^{1}_{t-1} {x^{2}_{t}} & \y x^{1}_{t-1} {x^{1}_{t-1}} & \y x^{1}_{t-1} {x^{2}_{t-1}} & \yy x^{1}_{t-1} {x^{1}_{t-2}} & \yy x^{1}_{t-1} {x^{2}_{t-2}}\\
                 \yy x^{2}_{t-1} {x^{1}_{t}} & \yy x^{2}_{t-1} {x^{2}_{t}} & \y x^{2}_{t-1} {x^{1}_{t-1}} & \y x^{2}_{t-1} {x^{2}_{t-1}} & \yy x^{2}_{t-1} {x^{1}_{t-2}} & \yy x^{2}_{t-1} {x^{2}_{t-2}}\\ \hline
                 \yyy x^{1}_{t-2} {x^{1}_{t}} & \yyy x^{1}_{t-2} {x^{2}_{t}} & \yy x^{1}_{t-2} {x^{1}_{t-1}} & \yy x^{1}_{t-2} {x^{2}_{t-1}} & \y x^{1}_{t-2} {x^{1}_{t-2}} & \y x^{1}_{t-2} {x^{2}_{t-2}}\\
                 \yyy x^{2}_{t-2} {x^{1}_{t}} & \yyy x^{2}_{t-2} {x^{2}_{t}} & \yy x^{2}_{t-2} {x^{1}_{t-1}} & \yy x^{2}_{t-2} {x^{2}_{t-1}} & \y x^{2}_{t-2} {x^{1}_{t-2}} & \y x^{2}_{t-2} {x^{2}_{t-2}}
            % \end{bmatrix}
            \end{array}
            \right]
            \right), \ \ \textrm{for} \ t \in \{ 1, \dots, n\}
        \end{split}
    \end{align*}
    \end{minipage}}
    \vspace{.75\baselineskip}
    \end{minipage}}
    \caption{Covariance matrix of the three-dimensional time-delay embedding $X'$ of a two-dimensional time series $X$. For simplicity, we here assume that $X'$ has zero mean.}
    \label{fig:cov-matrix}
\end{figure*}
\noindent The probability density functions $p_{I} = \mathcal{N(\mu_{I}, S_{I})}$ and $p_{\Omega} = \mathcal{N}(\mu_{\Omega}, S_{\Omega})$, for $\mu_{I}, \mu_{\Omega} \in \mathbb{R}^{d}$ and $S_{I}, S_{\Omega} \in \mathbb{R}^{d \times d}$, are approximated by a multivariate normal distribution, allowing for a closed-form solution of the KL divergence:
\begin{align}
    \begin{split}
        \mathcal{D}(p_{I}, p_{\Omega}) = &- \frac{1}{2} \Big( (\mu_{\Omega} - \mu_{I})^{\top} S^{-1}_{\Omega} (\mu_{\Omega} - \mu_{I})\\
        &+ \textrm{tr}(S^{-1}_{\Omega}S_{I}) + \log \frac{|S_{\Omega}|}{|S_{I}|} - d\Big).
    \end{split}
\end{align}

In order to account for the fact that the time series' samples are inter-dependent, the MDI algorithm applies time-delay embedding \cite{time-delay} as a pre-processing step. It creates a modified time series $X' := \{ x_{t}'\}_{t=1+(\kappa-1)\tau}^{n}$, $x_{t}' \in \mathbb{R}^{\kappa d}$, where each sample $x_{t}'$ includes attributes from previous $(\kappa - 1)\tau$ time steps as context:
\begin{equation}
    x_{t}' = 
    \begin{bmatrix}
        x_{t}^{\top} & x_{t - \tau}^{\top} & x_{t - 2\tau}^{\top} & \cdots & x_{t - (\kappa - 1)\tau}^{\top}
    \end{bmatrix}
    ^{\top},
\end{equation}

where $\kappa$ denotes the number of aggregated samples called the \textit{embedding dimension} and $\tau$ denotes the distance between two consecutive samples called the \textit{lag}. The underlying optimization problem for finding the most anomalous interval is:
$
\hat{I} = \underset{I \in \mathcal{I}}{\textrm{argmax}} \ \mathcal{D}(p_{I}, p_{\Omega(I)}).
$
However, according to the user's input, the MDI algorithm can output multiple, non-intersecting, anomalous intervals with the anomaly score ranked from the highest to the lowest. Furthermore, one can configure minimum and the maximum of the intervals to consider based on data-related domain knowledge in order to improve the algorithm's performance.

\section{Method}\label{method}

Let $I=[a,b)\cap \mathbb{N}$ be an anomalous interval of the multivariate time series $X$, detected by the MDI algorithm. Furthermore, let $x^{j}$, $j \in \mathcal{J}$, be the variables constituting $X$. To attribute the anomaly that occurred in $I$, we employ counterfactual reasoning in the sense that we intervene on a subset of variables of $X$ by making their values within $I$ more like the rest of the data within $\Omega = \{1, \dots, n\} \setminus I$. We then observe if the anomaly would still have occurred had this action been taken, i.e., had a certain subset of variables been less anomalous.

More specifically, we systematically modify a subset $X_{V}:=\{x^{k} \ | \ k \in V \}$ of variables with indices in $V \subset \mathcal{J}$, for $|V| \leq \left \lceil{\frac{d}{2}} \right \rceil$, by replacing their values within the anomalous interval $I$ with a random sample from the outside distribution $p_{\Omega}$. We assume that this is the nominal distribution of the data. We then re-apply the MDI scoring algorithm to the corrected interval to obtain a new anomaly score.
Suppose the new score is lower than before we modified a particular subset of variables $X_{V}$. Then, if all other subsets of equally many variables yield a higher anomaly score after the same procedure, we can conclude that the variables in $X_{V}$ contribute to the anomaly because of the type of the performed intervention that we call the \textit{counterfactual replacement}.

We will now define the counterfactual replacement of a subset of variables $X_V$ within the interval $I$ in more detail. As mentioned previously, it answers the question of how the variables of $X_V$ would look like inside the interval $I$ if they would have been governed by the same distribution $p_{\Omega}$ as the rest of the data within $\Omega$.
Clearly, the values of the variables from $X_V$ depend on the values of the variables from the complementary subset $X_{\overline{V}}$ in the same interval, which we do not replace.
An independent sample of $X_V$ would destroy these correlations and hence introduce another, potentially even stronger anomaly than the original one.
Therefore, the replacement must preserve the correlations with the non-replaced variables.
If not for the temporal dependencies between the consecutive time steps, we could simply draw $|I|$ independent samples from the conditional distribution $p_{\Omega}(\cdot \mid X_{\overline{V}})$ as a replacement for $X_V$. However, due to the time-delay embedding of the MDI algorithm, we must not only heed the inter-variable correlations but also the temporal dependencies in the data to obtain a valid counterfactual replacement.
Furthermore, the replacement should be conditional on the left and right context of the interval $I$, in order not to introduce an artificial anomaly due to a sudden break of temporal consistency.

In analogy to the principle of time-delay embedding, we model each time step to be replaced as an individual random variable.
Following the default distribution assumption of the MDI algorithm, we then use a multivariate normal distribution $\mathcal{N}(\mu, S)$, with $\mu \in \mathbb{R}^{d \ell}$, $S \in \mathbb{R}^{d \ell \times d \ell}$, for estimating the joint distribution of all subsets of the time series of length $\ell=b-a+2(\kappa-1)$.
This corresponds to the length of the anomalous interval plus the left and right context window, whose length depends on the time-delay embedding dimension $\kappa$ used by MDI for anomaly detection.
Obtaining the mean $\mu$ of this distribution is straightforward: Since it does not depend on the position of the samples in the time series, we can simply repeat the mean $\mu_\Omega$ of the nominal distribution $\ell$ times.

The covariance matrix $S$, in contrast, would be quite large and difficult to estimate robustly from limited data.
We observe, however, that in the limit $n \to \infty$ for infinitely long time series, $S$ converges to a symmetric block Toeplitz matrix, i.e., the blocks on each diagonal are constant.
The small illustrative example in Fig.~\ref{fig:cov-matrix} exhibits this property for a two-dimensional time series and $\ell=3$.
% whose covariance matrix is 
% $\textrm{cov}(X') = \left[
%     \begin{array}{ccc}
%         \color{blue!50}{A_{11}} & \color{green!50}{A_{12}} & \color{red!50}{A_{13}}\\
%         \color{green!50}{A_{21}} & \color{blue!50}{A_{22}} & \color{green!50}{A_{23}}\\
%         \color{red!50}{A_{31}} & \color{green!50}{A_{32}} & \color{blue!50}{A_{33}}
%     \end{array}\right]
% $.
The blocks having the same color code in Fig. \ref{fig:cov-matrix} have the same expectation value in the limit.
Therefore, we only need to estimate the first row of blocks from the available data, which fully defines the entire covariance matrix.
This reduces the computational complexity drastically from $\mathcal{O}(\ell^2 d^2)$ to $\mathcal{O}(\ell d^2)$.
To prevent the anomaly itself from influencing the estimate of the nominal distribution, we replace the interval $I$ with missing values before computing $\mu$ and $S$.

Finally, we draw a replacement for the selected variables $X_V$ within the anomalous interval $I$ from the conditional distribution $\mathcal{N}(\, \mu, S \mid X_{\overline{V}}, X_\mathrm{left}, X_\mathrm{right} \,)$.
We condition on the non-replaced variables $X_{\overline{V}}$ with $\overline{V} = \mathcal{J} \setminus V$ to maintain the inter-variable dependencies and on the left and right context $X_\mathrm{left} = [x_{a-(\kappa-1)}, \dotsc, x_{a-1}]^\top, X_\mathrm{right} = [x_{b}, \dotsc, x_{b+(\kappa-2)}]^\top$ to make the replacement connect smoothly to the surroundings of the anomalous interval.
An example of such a counterfactual replacement is shown in Fig.~\ref{fig:time-dependent-replacement}.

\section{Experiments}\label{experiments}

We applied our method to several well-understood extreme weather events and compared the attributed variables as the means of evaluation. In all the experiments, we used the time-delay embedding of dimension $\kappa=3$, the lag $\tau=1$ and the unbiased KL divergence $\textrm{U-KL}(p_{I}, p_{\Omega}):= 2 \cdot |I| \cdot \textrm{KL}(p_{I}, p_{\Omega})$ as rationalized by Barz et al. \cite{MDI}. Furthermore, we specified the size of the desired anomalous interval for each dataset individually. We applied our method to three different datasets of multivariate time series and one multivariate spatio-temporal dataset. The results in Tables \ref{fig:attribution-scores-hurricanes}-\ref{fig:attribution-scores-spatio-temporal} are obtained after averaging $10$ realizations of the in-distribution replacements of our attribution method scored by the KL divergence. In all our experiments, we discuss the attribution during and before the detected anomalous intervals to account for the lagged-effect of certain variables, i.e., whether the anomalous interval is an effect of a cause that started earlier in time than the detected interval.  

\subsection{Hurricanes}

\begin{table}[t]
	\centering
	\begin{tabular}{@{}c*{5}{c}@{}}
		\toprule
		\multicolumn{3}{c}{} & 10 days before & Detection\\
		\midrule
		\midrule
		\multicolumn{3}{c}{Anomaly score} & 104.69 & 371.44 \\
		\midrule
		\midrule
		SLP & & & 93.91 & 332.97 \\
		& W & & \yy \textbf{79.09} & 350.25 \\
		& & Hs & 105.58 & \yy \textbf{248.34}\\
		\midrule
		SLP & W & & \yy \textbf{57.11} & 333.4 \\
		SLP & & Hs & 107.32 & \yy \textbf{129.7} \\
		& W & Hs & 61.81 & 182.58 \\
		\bottomrule
	\end{tabular}
	\caption{Anomaly scores after counterfactual replacements on the Hurricanes dataset. The lowest scores for variable subsets of different cardinality are highlighted and shown in bold font.}
	\label{fig:attribution-scores-hurricanes}
\end{table}

We first apply our attribution approach to meteocean data obtained near the Bahamas in the Atlantic Sea from the National Data Buoy Center from the National Oceanic and Atmospheric Administration (NOAA) \cite{oceans-data}. It encompasses six months of hourly measurements of significant wave height (Hs), wind speed (W), and sea level pressure (SLP), from June 2012 until November 2012. During this time, the Atlantic hurricane season was especially active with the occurrence of $19$ tropical cyclones ($\textrm{W} > 52$ km/h), $10$ of which became hurricanes ($\textrm{W} > 64$ km/h).

We set the size of the intervals to be searched for by the MDI algorithm between $24$ and $120$ hours. In Fig. \ref{fig:time-dependent-replacement}, we show the interval replacement of the Hs variable within the detected anomalous interval. Table \ref{fig:attribution-scores-hurricanes} shows our attribution results during the detected hurricane, as well as ten days before the event. The variable names indicate which subset of variables was counterfactually replaced within the anomalous interval $I$. When only one variable at a time is replaced within $I$, we observe that the anomaly score improves the most in the case of the wave height. This means that the MDI algorithm mostly detects this interval as anomalous due to the anomaly in Hs. However, when inspecting the interval ten days prior to the occurrence of the detected hurricane, we notice that modifying the wind speed during that period lowers the anomaly score. When two-variable subsets are counterfactually replaced within $I$, we note that W helps lower the anomaly score even further during the event and can therefore be attributed to the MDI's detection together with Hs. However, ten days prior to the event, SLP along with W reduces the anomaly score the most when their values within $I$ are replaced by the in-distribution samples. This could imply that SLP is the cause of the hurricane itself. As expected, when we repair all three variables, the anomaly score drastically decreases. For this reason, in all the experiments we assume that at most $\frac{d}{2}$ variables are responsible for the event. Otherwise, by changing more variables in the interval $I$, we drastically change the settings of the event which could result in an inaccurate attribution.

\subsection{Multivariate Ecological Time Series}
\begin{table}[t]
    \centering
    \tiny
	\begin{tabular}{@{}c*{8}{c}@{}}
		\toprule
		\multicolumn{6}{c}{DE-Hai} & One month before & Detection\\
		\midrule
		\midrule
		\multicolumn{6}{c}{Anomaly score} & 21.63 & 142.57\\
		\midrule
		\midrule
		T$_{\textrm{air}}$ & & & & & & 20.54 & 140.99\\
		& NEE & & & & & 20.79 & 133.89\\
		& & PPT & & & & \yy \textbf{19.7} & 118.99\\
		& & & VPD & & & 21.16 & \yy \textbf{51.17}\\
		& & & & LE & & 20.72 & 113.45\\
		& & & & & H & 20.52 & 141.03\\
		\midrule
		T$_{\textrm{air}}$ & NEE & & & & & 19.1 & 135.04\\
		T$_{\textrm{air}}$ & & PPT & & & & 19.64 & 115.53\\
		T$_{\textrm{air}}$ & & & VPD & & & 18.8 & 37.39\\
		T$_{\textrm{air}}$ & & & & LE & & 19.59 & 99.51\\
		T$_{\textrm{air}}$ & & & & & H & 19.63 & 136.95\\
		& NEE & PPT & & & & 19.4 & 112.61\\
		& NEE & & VPD & & & 20.34 & 50.27\\
		& NEE & & & LE & & \yy \textbf{18.08} & 109.38\\
		& NEE & & & & H & 20.75 & 134.87\\
		& & PPT & VPD & & & 20.26 & \yy \textbf{30.13}\\
		& & PPT & & LE & & 18.98 & 90.85\\
		& & PPT & & & H & 18.95 & 118.53\\
		& & & VPD & LE & & 20.34 & 46.29\\
		& & & VPD & & H & 20.01 & 44.63\\
		& & & & LE & H & 20.38 & 109.95\\
		\midrule
		T$_{\textrm{air}}$ & NEE & PPT & & & & 18.01 & 111.09\\
		T$_{\textrm{air}}$ & NEE &  & VPD & & & 18.37 & 37.12\\
		T$_{\textrm{air}}$ & NEE & & & LE & & \yy \textbf{16.07} & 87.58\\
		T$_{\textrm{air}}$ & NEE & & & & H & 20.39 & 135.07\\
		T$_{\textrm{air}}$ & & PPT & VPD & & & 18.35 & \yy \textbf{18.6}\\
		T$_{\textrm{air}}$ & & PPT & & LE & & 17.06 & 79.11\\
		T$_{\textrm{air}}$ & & PPT & & & H & 17.58 & 113.83\\
		T$_{\textrm{air}}$ & & & VPD & LE & & 16.74 & 36.36\\
		T$_{\textrm{air}}$ & & & VPD & & H & 19.07 & 34.45\\
		T$_{\textrm{air}}$ & & & & LE & H & 18.07 & 99.87\\
		& NEE & PPT & VPD & & & 18.37 & 31.24\\
		& NEE & PPT & & LE & & 16.98 & 85.75\\
		& NEE & PPT & & & H & 19.25 & 114.6\\
		& NEE & & VPD & LE & & 18.39 & 44.17\\
		& NEE & & VPD & & H & 19.26 & 43.34\\
		& NEE & & & LE & H & 17.74 & 102.92\\
		& & PPT & VPD & LE & & 20.35 & 26.2\\
		& & PPT & VPD & & H & 18.1 & 24.75\\
		& & PPT & & LE & H & 18.4 & 85.02\\
		& & & VPD & LE & H & 19.32 & 42.09\\
		\bottomrule
	\end{tabular}
	\caption{Anomaly scores after the counterfactual replacements one month before and during the detected anomaly from July and August 2003 on the German site DE-Hai. The lowest scores for variable subsets of different cardinality are highlighted and shown in bold font.}
	\label{fig:attribution-scores-DE-Hai-te3}
\end{table}
We perform anomaly attribution in multivariate time series of the certain ecological sites of the FLUXNET \cite{FLUXNET} dataset. Particularly, we analyze the causes of the European heatwave in the Summer of $2003$ in Germany and France, by using the data from the site DE-Hai and FR-Pue, respectively.

\subsubsection{German Heatwave Attribution}
\begin{figure}[t]
    \centering
    \includegraphics[width=1\columnwidth]{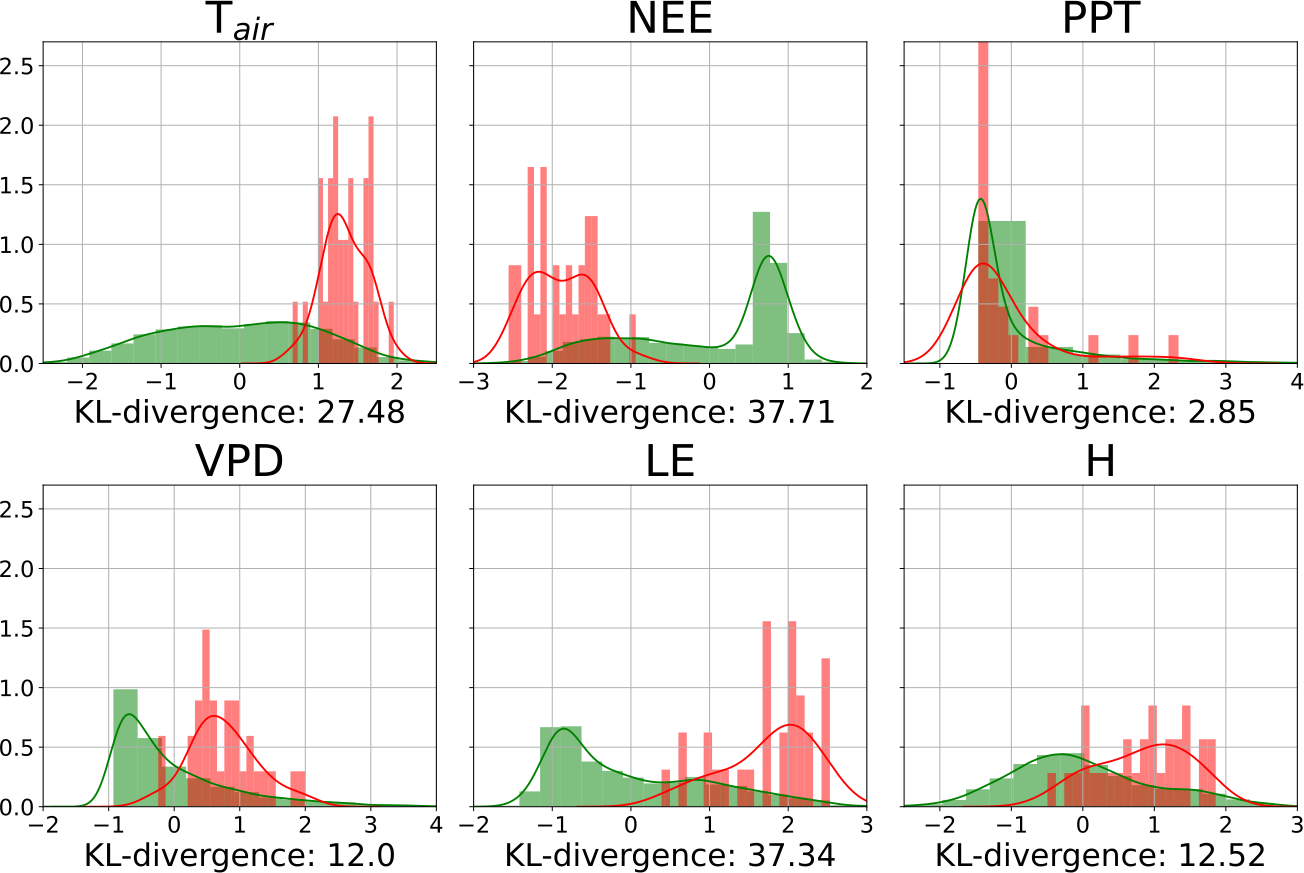}
    \caption{Attribution scheme of the German heatwave. The histogram of the entire variable is shown in green, whereas the histogram of the variable within the anomalous interval is shown in red. The KL-divergence results of the univariate comparison of the red and green histograms are shown individually for each variable under the $x$-axis.}
    \label{fig:DE-histograms}
\end{figure}
\begin{table}[t]
    \centering
    \tiny
    \begin{tabular}{@{}c*{8}{c}@{}}
    	\toprule
    	\multicolumn{6}{c}{FR-Pue} & One month before & Detection\\
    	\midrule
    	\midrule
    	\multicolumn{6}{c}{Anomaly score} & 20.93 & 123.38\\
    	\midrule
    	\midrule
    	T$_{\textrm{air}}$ & & & & & & 18.9 & 113.78\\
    	& NEE & & & & & \yy \textbf{17.61} & 116.88\\
    	& & PPT & & & & 20.56 & 83.61\\
    	& & & VPD & & & 18.77 & \yy \textbf{83.54}\\
    	& & & & LE & & 17.9 & 120.91\\
    	& & & & & H & 20.88 & 121.26\\
    	\midrule
    	T$_{\textrm{air}}$ & NEE & & & & & 16.37 & 110.21\\
    	T$_{\textrm{air}}$ & & PPT & & & & 17.83 & 75.35\\
    	T$_{\textrm{air}}$ & & & VPD & & & 17.79 & 74.02\\
    	T$_{\textrm{air}}$ & & & & LE & & \yy \textbf{14.5} & 109.76\\
    	T$_{\textrm{air}}$ & & & & & H & 18.61 & 110.51\\
    	& NEE & PPT & & & & 18.66 & 77.07\\
    	& NEE & & VPD & & & 17.19 & 73.74\\
    	& NEE & & & LE & & 14.99 & 97.43\\
    	& NEE & & & & H & 18.44 & 115.41\\
    	& & PPT & VPD & & & 17.56 & \yy \textbf{42.88}\\
    	& & PPT & & LE & & 15.99 & 81.86\\
    	& & PPT & & & H & 20.69 & 82.5\\
    	& & & VPD & LE & & 16.88 & 86.95\\
    	& & & VPD & & H & 19.66 & 79.58\\
    	& & & & LE & H & 19.05 & 124.36\\
    	\midrule
    	T$_{\textrm{air}}$ & NEE & PPT & & & & 18.24 & 69.55\\
    	T$_{\textrm{air}}$ & NEE &  & VPD & & & 17.35 & 67.88\\
    	T$_{\textrm{air}}$ & NEE & & & LE & & 14.43 & 88.04\\
    	T$_{\textrm{air}}$ & NEE & & & & H & 18.21 & 107.98\\
    	T$_{\textrm{air}}$ & & PPT & VPD & & & 14.73 & \yy \textbf{35.28}\\
    	T$_{\textrm{air}}$ & & PPT & & LE & & 13.94 & 69.98\\
    	T$_{\textrm{air}}$ & & PPT & & & H & 18.27 & 70.36\\
    	T$_{\textrm{air}}$ & & & VPD & LE & & \yy \textbf{13.84} & 82.69\\
    	T$_{\textrm{air}}$ & & & VPD & & H & 18.17 & 61.46\\
    	T$_{\textrm{air}}$ & & & & LE & H & 14.7 & 106.02\\
    	& NEE & PPT & VPD & & & 16.51 & 35.58\\
    	& NEE & PPT & & LE & & 14.46 & 55.12\\
    	& NEE & PPT & & & H & 18.71 & 75.73\\
    	& NEE & & VPD & LE & & 14.8 & 65.91\\
    	& NEE & & VPD & & H & 16.91 & 71.76\\
    	& NEE & & & LE & H & 16.55 & 96.49\\
    	& & PPT & VPD & LE & & 15.5 & 47.42\\
    	& & PPT & VPD & & H & 18.66 & 39.89\\
    	& & PPT & & LE & H & 17.88 & 78.73\\
    	& & & VPD & LE & H & 17.52 & 82.83\\
    	\bottomrule
    \end{tabular}
    \caption{Anomaly scores after counterfactual replacements one month before and during the detected anomaly from mid-July to mid-August 2003 on the French site FR-Pue. The lowest scores for variable subsets of different cardinality are highlighted and shown in bold font.}
    \label{fig:attribution-scores-FR-Pue-te3}
\end{table}
To investigate the extreme precipitation event, namely the lack of rain that occurred in Germany in 2003 during the European heatwave, we take a closer look at the FLUXNET site DE-Hai, located in the Hainich National Park. We use the daily measurements of the air temperature (T$_{\textrm{air}}$), net ecosystem exchange (NEE), precipitation (PPT), vapor pressure deficit (VPD), latent heat flux (LE), and sensible heat (H), as suggested by Krich et al. \cite{Christopher}. The data span from the year 2000 until 2009. We first apply the MDI method to detect the two-month-long anomalous interval, which coincidentally happens to occur exactly from July until the end of August 2003, when the European heatwave had occurred as well. Then we perform systematic in-distribution replacements within the detected interval for subsets of variables with at most three variables and re-score it. The results thereof are shown in Table \ref{fig:attribution-scores-DE-Hai-te3}. When considering singleton variable subsets, we note that the most responsible variable for the MDI's detection of this interval is VPD. When replacing an anomalous segment of two variables at a time, we find that this extreme event can be attributed to PPT, along with VPD, which is in line with Krich et al. \cite{Christopher}. Moreover, when we replace three variables at a time within the detected interval, we observe that T$_{\textrm{air}}$ in addition to PPT and VPD reduces the anomaly score the most. This is also a well-supported attribution when studying an extreme event such as a heatwave. In addition, we inspected the one-month-long interval before the detected event and observed that the most anomalous singleton consists of PPT, the most anomalous two-variable subset consists of NEE and LE and adding T$_{\textrm{air}}$ to the latter two variables seemed to have been the most dissimilar to the rest of the data one month prior to the event. This may have contributed to the occurrence of the said heatwave.

To emphasize the importance of considering inter-variable correlations in multivariate time series during attribution, we furthermore conduct a simple baseline attribution experiment: Each variable is considered in isolation and the anomaly scores during the time frame in question are compared. Fig. \ref{fig:DE-histograms} shows that such a purely univariate approach would arrive at the false conclusion of NEE and LE being the most anomalous variables, whereas we know from our multivariate attribution scheme that VPD and PPT actually account for the anomaly.

\subsubsection{French Heatwave Attribution}

The FLUXNET site FR-Pue which we used to inspect the European heatwave in France, consists of the $14$ years of daily measurements of the ecological variables T$_{\textrm{air}}$, NEE, PPT, VPD, LE, and H, in the period from $2001$ until $2014$. In Table \ref{fig:attribution-scores-FR-Pue-te3}, we see the results of counterfactually replacing different subsets of variables within the anomalous interval, while carefully minding the inter-temporal relations. We note that the anomaly was detected from mid-July until mid-August of 2003, whereas the heatwave was documented through the entire July and August of the same year. When counterfactually modifying only one variable at the time, we see in Table \ref{fig:attribution-scores-FR-Pue-te3} that VPD lowers the anomaly score the most, meaning that it also contributes to the anomaly the most. In the case of substituting the anomalous segments of two variables at the time, PPT and VPD yield the lowest anomaly score, and we attribute this heatwave to them,. This is similar to the attribution of the same event on the German site DE-Hai. When replacing three variables' values with the in-distribution ones within the detected interval, we note that $T_{\textrm{air}}$ contributes to lowering the anomaly score in addition to PPT and VPD. This is in line with Shadaydeh et al. \cite{Maha-GCPR} who also attributed this event to $T_{\textrm{air}}$ and VPD. Furthermore, as for the German site, we analyzed the one-month-long interval prior to the detected anomaly and noted that in this case NEE was the most anomalous, potentially contributing to the occurrence of the heatwave in question. Two-variable replacements a month before the detection indicated that the event could have been caused by abnormalities of $T_{\textrm{air}}$ and LE. Moreover, when simultaneously replacing three variables with the in-distribution values one month prior to the detected anomaly, we see that T$_{\textrm{air}}$, VPD and LE yield the lowest anomaly score.

\subsection{Spatio-Temporal Data}

\begin{table}[t]
    \centering
    \begin{tabular}{@{}c*{5}{c}@{}}
		\toprule
		\multicolumn{3}{c}{Spatio-temporal data} & 10 days before & Detection\\
		\midrule
		\midrule
		\multicolumn{3}{c}{Anomaly score} & $4.67$ & $6.11$ \\
		\midrule
		\midrule
		SLP & & & $200$ & $340$\\
		& TS & & $60$ & \yy $\textbf{2.69}$ \\
		& & W & $90$ & $380$\\
		\midrule
		SLP & TS & & \yy $\textbf{2.6}$ & $3.1$\\
		SLP & & W & $10$ & $50$\\
		& TS & W & $6.71$ & \yy $\textbf{2.95}$\\
		\bottomrule
	\end{tabular}
	\caption{Anomaly scores after counterfactual replacements 10 days before and during the 12-day-long detected extreme weather event in spatio-temporal data. The lowest scores lower than the anomaly score, for variable subsets of different cardinality are highlighted and shown in bold font. All values are given in units of $10^{5}$.}
	\label{fig:attribution-scores-spatio-temporal}
\end{table}
For the experiments on spatio-temporal data, we use the data provided by Racah et al. \cite{marine-data}. It consists of spatio-temporal variables of the weather conditions on Earth and boxes outlying the hurricanes along with a class label. Four samples per day are available for a period of $365$ days.
For our anomaly attribution analysis, we choose the variables of sea level pressure (SLP), surface temperature (TS), and wind speed (W), similarly to the variables of the purely temporal Hurricanes dataset.

We apply our method pixel-wise to a region of $220 \times 220$ pixels, where we know a hurricane has occurred, to achieve computational tractability. We preset the spatial size of the detection for the spatio-temporal MDI algorithm to regions of $50 \times 50$ pixels, lasting $12$ days. In Table \ref{fig:attribution-scores-spatio-temporal}, we show our method's results ten days before and during the detected extreme event. The detected event can be attributed to TS as counterfactually replacing that variable within the detection interval reduces the anomaly score the most. Interestingly, when we counterfactually replace two variables at a time within the corresponding time frames, in addition to TS, ten days before the anomaly, SLP influences the anomaly score the strongest. In contrast, during the detection itself, the anomaly score is influenced the most when replacing W and TS.

\section{Conclusion}\label{conclusion}

We proposed a novel anomaly attribution approach for multivariate temporal and spatio-temporal data based on the MDI method for anomaly detection and the counterfactual replacements of variables within the anomalous intervals. This counterfactual replacement answers the question of would the anomaly still have occurred, had a subset of variables been more similar to the data outside of that interval. The main benefit of our attribution method is that it can be applied to any multivariate time series data, regardless of potential outliers or missing values, since it replaces the entire anomalous interval with in-distribution samples and re-scores it. It also considers the correlations both among variables and across time. It is thus also suitable for the task of data repairing. Furthermore, it is one of the few methods for multivariate anomaly attribution in time series, and provides interpretation of the MDI method's anomaly detections. We demonstrated the use of our method on multiple multivariate time series datasets and one multivariate spatio-temporal dataset. By obtaining attributions for well-documented extreme events whose causes are well-understood, we confirmed the correctness of our method. Therefore, we conclude that our approach is a fast and accurate tool to help domain experts in understanding the possible causes of anomalous events in multivariate time series. Moreover, the current work can be further exploited towards the classification of different anomalous events based on the presented attribution.

\bibliographystyle{ieeetr}
\bibliography{bibliography-AAK}

\end{document}